\documentclass[journal]{IEEEtran}
\usepackage[colorlinks,urlcolor=blue,linkcolor=blue,citecolor=blue]{hyperref}
\usepackage[table]{xcolor}
\usepackage{tabularx}
\usepackage{amssymb}
\usepackage{color,array}
\usepackage{booktabs}
\usepackage{graphicx}
\usepackage{pifont}   
\usepackage{amsmath} 
\usepackage[utf8]{inputenc} 
\usepackage{array} 
\usepackage{multirow}
\usepackage[T1]{fontenc}
\definecolor{myGreen}{RGB}{34,139,34} 
\definecolor{myRed}{RGB}{220,20,60}   
\definecolor{rowgray}{gray}{0.93} 

\usepackage{graphicx}

\usepackage{xurl}      
\usepackage{fvextra}   

\usepackage[most]{tcolorbox}
\tcbuselibrary{skins,breakable,listings}

\newtcolorbox{promptbox}[1]{%
  enhanced,
  breakable,
  colback=white,
  colframe=gray!70,
  colbacktitle=gray!70,
  coltitle=black,
  title={#1},
  fonttitle=\bfseries,
  fontupper=\footnotesize,
  boxrule=0.8pt,
  arc=2mm,
  left=6pt,right=6pt,top=6pt,bottom=6pt,
  boxsep=4pt
}

\newtcblisting{jsonbox}{%
  enhanced,
  breakable,
  colback=gray!5,
  colframe=gray!40,
  boxrule=0.4pt,
  arc=1.5mm,
  left=4pt,right=4pt,top=4pt,bottom=4pt,
  listing only,
  listing options={
    basicstyle=\ttfamily\footnotesize,
    breaklines=true,
    breakatwhitespace=false,
    columns=fullflexible,
    keepspaces=true,
    showstringspaces=false
  }
}

\begin{document}

\title{OmniMER: Auxiliary-Enhanced LLM Adaptation for Indonesian Multimodal Emotion Recognition} 
\author{Xueming Yan, \IEEEmembership{Senior Member, IEEE}, Boyan Xu, Yaochu Jin, \IEEEmembership{Fellow, IEEE}, \\ Lixian Xiao,  Wenlong Ye, Runyang Cai,  Zeqi Zheng, Jingfa Liu, \\ Aimin Yang and Yongduan~Song,~\IEEEmembership{Fellow,~IEEE}
\thanks{Manuscript received XX Dec 2025; revised XX, XXXX.
This work was supported in part by National Natural Science Foundation of China (62576116, 62136003),  International Collaboration Fund for Creative Research Teams (ICFCRT) of NSFC (W2441019), the Major Research Plan of the National Natural Science Foundation of China (92570116), the Major Program of the Natural Science Foundation of Zhejiang Province (D25F020001), the Young Researcher Startup Program of Guangdong-Hong Kong-Macau Center for Applied Mathematics (2025A1515060002) and the Guangdong Philosophy and Social Science Foundation Regular Project (GD25YGG26). \textit{(Corresponding author: Yaochu Jin)}
}
\thanks{X. Yan, R. Cai and J. Liu is with the School of Information Science and Technology, Guangdong University of Foreign Studies, Guangzhou, 510006. Email: yanxm@gdufs.edu.cn; 20231010047@mail.gdufs.edu.cn; jfliu@gdufs.edu.cn}  
\thanks{B. Xu and W. Ye is with  the School of Computer Science, Guangdong University of Technology, Guangzhou, 510006. Email: hpakyim@gmail.com; ywlaaron@gmail.com}
\thanks{L. Xiao is with Faculty of Asian Languages and Cultures, Guangdong University of Foreign Studies, Guangzhou, China. Email: 200110732@oamail.gdufs.edu.cn}
\thanks{Y. Jin and Z. Zheng is with the School of Engineering, Westlake University, Hangzhou 310030, China. Email:jinyaochu@westlake.edu.cn; zhengzeqi@westlake.edu.cn}  %
\thanks{A. Yang is with School of Computer Science and Intelligence Education, Lingnan Normal University, Zhanjiang, 524000, China. Email: amyang@gdut.edu.cn.}
\thanks{Y. Song is with the School of Automation, Chongqing University, Chongqing, 400044, China. Email: ydsong@cqu.edu.cn.}
\thanks{X. Yan and B. Xu contributed equally to this work.}
}
\markboth{Journal of \LaTeX\ Class Files,~Vol.~xx, No.~x, August~20xx}%
{Shell \MakeLowercase{\textit{et al.}}: Bare Demo of IEEEtran.cls for IEEE Journals}

\maketitle

\begin{abstract}
Indonesian, spoken by over 200 million people, remains understudied in multimodal emotion recognition research despite its dominant presence on Southeast Asian social media platforms. To fill the gap, we introduce IndoMER, the first multimodal emotion recognition benchmark for Indonesian, which comprises 1,944 video segments from 203 speakers with temporally aligned text, audio, and visual annotations across seven emotion categories. The dataset poses realistic challenges for emotion recognition, including cross-modal inconsistency and long-tailed class distributions, which can be partly attributed to Indonesian culture. To address these challenges, we propose OmniMER, a multimodal adaptation framework built upon Qwen2.5-Omni that enhances emotion recognition through three auxiliary modality-specific perception tasks: emotion keyword extraction for text, facial expression analysis for video, and prosody analysis for audio. These auxiliary tasks help the model identify emotion-relevant cues in each modality before fusion, reducing the reliance on spurious correlations in low-resource settings. Experiments on IndoMER show that OmniMER achieves 0.582 Macro-F1 on sentiment classification and 0.454 on emotion recognition, outperforming the baseline model by 7.6 and 22.1 absolute points, respectively. Cross-lingual evaluation on the Chinese CH-SIMS dataset further demonstrates the generalizability of the proposed framework. The dataset and code are publicly available. (https://github.com/yanxm01/INDOMER)
\end{abstract}


\section{Introduction}
\IEEEPARstart{E}{motion} recognition is central to affective computing and supports a wide range of applications, from mental health monitoring \cite{guo2024development} to social media content moderation 
\cite{khowaja2024depression,liu2025hierarchical}. As multimodal data becomes increasingly available, multimodal emotion recognition (MER), which integrates text, audio, and visual signals, has gained significant attention \cite{sun2023efficient,wu2025comprehensive}. Text conveys explicit semantic content, audio captures prosodic and vocal cues, and video encodes facial expressions and head movements. By combining these complementary modalities, MER systems capture the complexity of human affect more comprehensively 
than unimodal approaches \cite{deng2021survey,latif2021survey,karnati2023understanding}. However, the effectiveness of MER depends heavily on large amounts of high-quality annotated data, which remains untrue for Indonesian.

Despite significant progress, existing MER research remains heavily English-centric. Benchmark datasets such as IEMOCAP \cite{busso2008iemocap}, CMU-MOSEI \cite{zadeh2018multimodal}, and MELD \cite{poria2018meld} have driven methodological advances, yet they predominantly feature English speakers in acted or scripted 
scenarios, limiting their capacity to generalize to spontaneous, cross-cultural emotional expressions. Although CMU-MOSEAS \cite{zadeh2020cmu} extends coverage to Spanish, Portuguese, German, and French, these remain relatively high-resource European languages. A critical gap persists for languages in Asia, Africa, and 
other underrepresented regions, where both annotated multimodal data and pretrained linguistic resources are scarce (Table~\ref{dataset_comparison}).

Among these underrepresented languages, Indonesian stands out as a particularly significant case. With over 200 million speakers \cite{pepinsky2024urbanization}, Indonesian is the fourth most widely spoken language globally and dominates social media platforms across Southeast Asia \cite{al2025indonesia}. Despite 
this significant presence, Indonesian remains severely underrepresented in multimodal emotion research. To our knowledge, no publicly available multimodal emotion recognition dataset exists for Indonesian. This absence not only constrains the development of culturally appropriate emotion-aware systems for Indonesian speakers, but also limits cross-lingual evaluation and transfer learning research in genuinely low-resource settings \cite{girija2023analysis}.

Beyond data scarcity, MER in low-resource settings faces two compounding challenges. The first is cross-modal inconsistency. Emotional signals across modalities often conflict due to cultural communication norms \cite{wang2023incomplete,yan2022multimodal}. Indonesian speakers, influenced by values of social harmony (\textit{Rukun}) and politeness (\textit{Sopan Santun}), often use indirect or neutral language in text while displaying stronger emotional cues through vocal prosody or facial expressions \cite{lindquist2022cultural}. 
Such modality conflicts make it difficult for models to learn consistent emotional representations.
The second challenge is long-tailed class distributions. Natural emotional communication skews heavily towards neutral and positive expressions, while emotions such as fear and disgust are rare \cite{meng2024deep}. This imbalance biases standard training procedures towards majority classes and amplifies spurious correlations, where models rely on dataset artifacts rather than genuine emotional signals \cite{maheronnaghsh2024robustness}. With limited training data, these issues are further exacerbated, as models tend to overfit superficial patterns rather than learn robust multimodal representations \cite{myakala2023bridging,kim2024causal}.

Recent advances in omni-modal large language models (LLMs) offer a promising direction for addressing these challenges. Models such as GPT-4o \cite{islam2025gpt}, Gemini \cite{team2023gemini}, and Qwen2.5-Omni \cite{xu2025qwen2} unify text, audio, and visual understanding within a single architecture, demonstrating 
strong cross-modal perception capabilities. Their pretrained knowledge may help mitigate the limitations of small-scale datasets. However, directly applying omni-modal LLMs to emotion recognition without task-specific 
adaptation often yields suboptimal results, since theses models may rely on superficial patterns rather than genuine emotional cues \cite{lian2024merbench}. This motivates the need for targeted adaptation strategies that strengthen emotional grounding within each modality before multimodal fusion.

\begin{table*}[t]
\centering
\small 
\setlength{\tabcolsep}{10pt} 

\caption{Comparison of IndoMER with other publicly available multimodal datasets. Note: All listed datasets contain textual ($l$), audio ($a$), and \textbf{visual} ($v$) modalities. \ding{51} denotes availability, while \ding{55} denotes unavailability.}
\label{dataset_comparison}

\begin{tabular}{@{} l r r c c l @{}}
\toprule
\textbf{Dataset} & \textbf{Samples} & \textbf{Speakers} & \textbf{Sentiment} & \textbf{Emotion} & \textbf{Language} \\
\midrule
CMU-MOSEAS \cite{zadeh2020cmu} & 40,000 & 1,645 & \textcolor{myGreen}{\ding{51}} & \textcolor{myGreen}{\ding{51}} (Continuous) & 
\includegraphics[height=1.2ex]{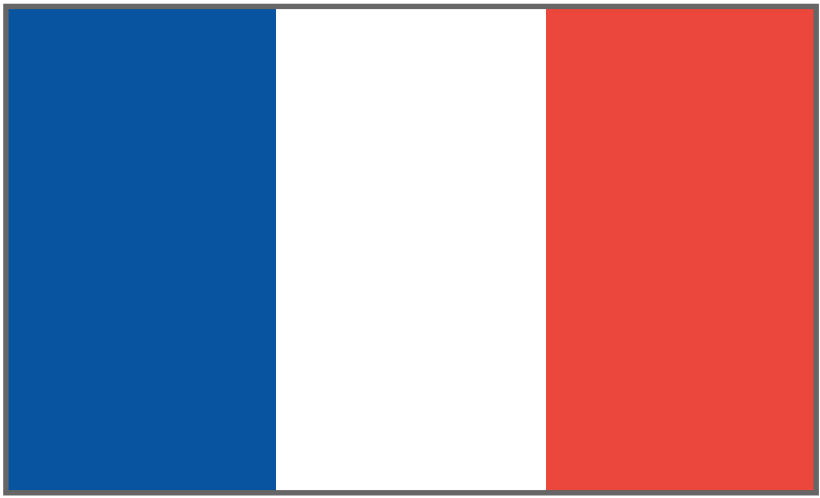} 
\includegraphics[height=1.2ex]{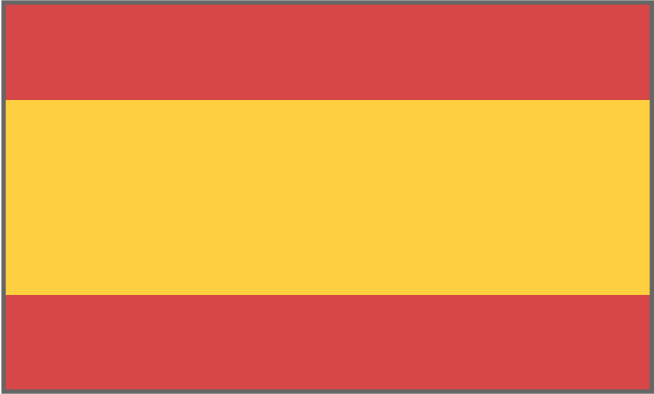} 
\includegraphics[height=1.2ex]{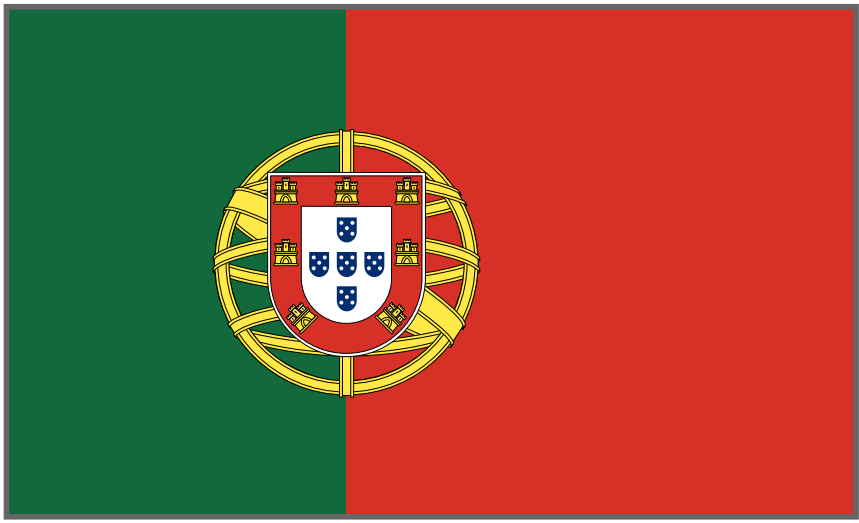} 
\includegraphics[height=1.2ex]{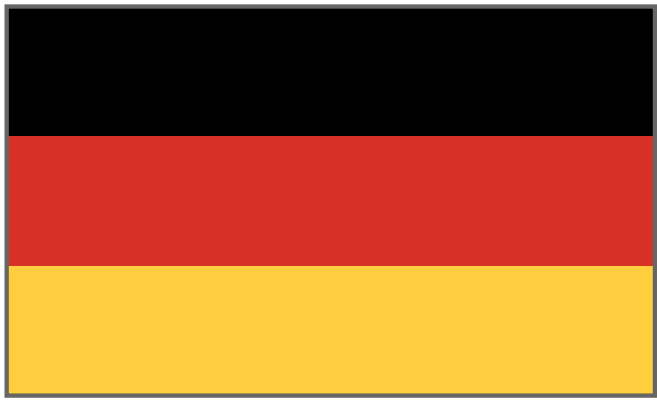} 
(Multi-lingual) \\

M3ED \cite{zhao2022m3ed} & 24,400 & 500 & \textcolor{myRed}{\ding{55}} & \textcolor{myGreen}{\ding{51}} (Discrete) & \includegraphics[height=1.2ex]{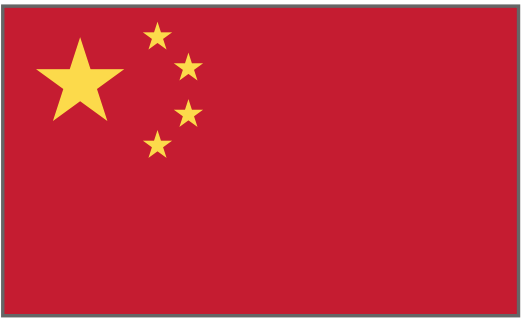} Chinese \\
MER-MULTI \cite{lian2023mer} & 3,784 & - & \textcolor{myGreen}{\ding{51}} & \textcolor{myGreen}{\ding{51}} (Discrete) & \includegraphics[height=1.2ex]{pdf/china.png} Chinese \\
CH-SIMS \cite{liu2022make} & 2,281 & 474 & \textcolor{myGreen}{\ding{51}} & \textcolor{myRed}{\ding{55}} & \includegraphics[height=1.2ex]{pdf/china.png} Chinese \\

UR-FUNNY \cite{hasan2019ur} & 16,514 & 1,741 & \textcolor{myRed}{\ding{55}} & \textcolor{myGreen}{\ding{51}} (Humor) & \includegraphics[height=1.2ex]{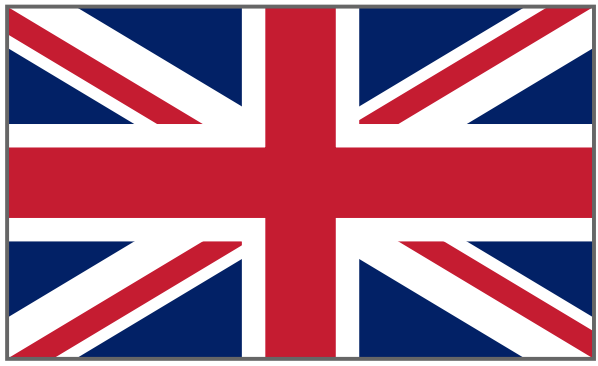} English \\
CMU-MOSEI \cite{zadeh2016mosi} & 23,453 & 1,000 & \textcolor{myGreen}{\ding{51}} & \textcolor{myGreen}{\ding{51}} (Continuous) & \includegraphics[height=1.2ex]{pdf/UnitedKingdom.png} English \\
MELD \cite{poria2018meld} & 13,708 & 260 & \textcolor{myGreen}{\ding{51}} & \textcolor{myGreen}{\ding{51}} (Discrete) & \includegraphics[height=1.2ex]{pdf/UnitedKingdom.png} English \\
EmoryNLP \cite{zahiri2018emotion} & 12,606 & - & \textcolor{myGreen}{\ding{51}} & \textcolor{myGreen}{\ding{51}} (Discrete) & \includegraphics[height=1.2ex]{pdf/UnitedKingdom.png} English \\
IEMOCAP \cite{busso2008iemocap} & 10,000 & 10 & \textcolor{myGreen}{\ding{51}} & \textcolor{myGreen}{\ding{51}} (Discrete) & \includegraphics[height=1.2ex]{pdf/UnitedKingdom.png} English \\
CMU-MOSI \cite{zadeh2016mosi} & 2,199 & 89 & \textcolor{myGreen}{\ding{51}} & \textcolor{myRed}{\ding{55}} & \includegraphics[height=1.2ex]{pdf/UnitedKingdom.png} English \\
MUStARD \cite{castro2019towards} & 690 & 23 & \textcolor{myGreen}{\ding{51}} & \textcolor{myRed}{\ding{55}} & \includegraphics[height=1.2ex]{pdf/UnitedKingdom.png} English \\
MOUD \cite{perez2013utterance} & 498 & 80 & \textcolor{myGreen}{\ding{51}} & \textcolor{myRed}{\ding{55}} & \includegraphics[height=1.2ex]{pdf/Spanish.png} Spanish \\

\midrule 
\rowcolor{rowgray} 
\textbf{IndoMER (Ours)} & \textbf{1,944} & \textbf{203} & \textbf{\textcolor{myGreen}{\ding{51}}} & \textbf{\textcolor{myGreen}{\ding{51}} (Discrete)} & \textbf{\includegraphics[height=1.2ex]{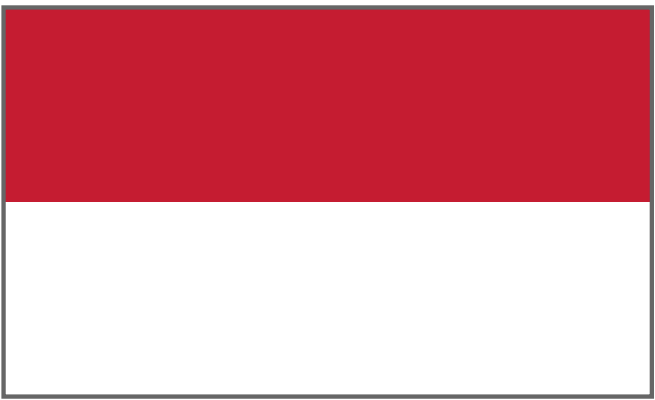} Indonesian} \\
\bottomrule
\end{tabular}
\end{table*}

To address these challenges, we introduce IndoMER, the first multimodal emotion recognition benchmark for Indonesian, and propose OmniMER, a framework that adapts omni-modal LLMs through auxiliary modality-specific 
perception tasks. Unlike conventional fusion approaches that directly combine multimodal features, OmniMER first strengthens unimodal emotional representations through three auxiliary tasks: emotion keyword extraction 
for text, facial expression analysis for video, and prosody analysis for audio. By grounding emotional inference in modality-specific evidence before fusion, OmniMER reduces reliance on spurious correlations and 
improves robustness under cross-modal conflict and class imbalance. Table~\ref{dataset_comparison} summarizes the position of IndoMER within the landscape of existing multimodal emotion datasets.

Our contributions are summarized as follows:
\begin{itemize}
    \item We introduce IndoMER, the first multimodal emotion recognition benchmark for Indonesian, which comprises 1,944 video segments from 203 speakers with temporally aligned text, audio and visual annotations across seven emotion categories.

    \item We propose OmniMER, a multimodal LLM adaptation framework that leverages three auxiliary modality-specific perception tasks to strengthen unimodal emotional grounding prior to fusion, improving robustness in low-resource settings.

    \item Through extensive experiments on IndoMER, we demonstrate that OmniMER outperforms the baseline model by 7.6 and 22.1 absolute points, respectively, on Macro-F1 for sentiment classification and emotion recognition, 
    with auxiliary tasks providing notable gains for minority emotion categories. Cross-lingual evaluations on CH-SIMS further validate the generalizability of the framework. 
\end{itemize}

The remainder of this paper is organized as follows. Section II describes the related work, Section III introduces the IndoMER dataset, Section IV introduces the proposed OmniMER framework, Section V reports experimental results, and Section VI concludes the study.

\section{Related Work}
This section reviews related work, including multimodal emotion recognition, low-resource learning, and advances in multimodal large language models with auxiliary-task supervision.

\subsection{Multimodal Emotion Recognition}
Multimodal emotion recognition has evolved significantly, driven by foundational benchmarks such as IEMOCAP~\cite{busso2008iemocap}, CMU-MOSI~\cite{zadeh2016mosi}, CMU-MOSEI~\cite{zadeh2018multimodal}, and MELD~\cite{poria2018meld}. These datasets enabled the development of diverse fusion strategies. Early approaches include early fusion, which concatenates features at the input level~\cite{poria2015deep}, and late fusion, which combines modality-specific predictions at the output layer~\cite{poria2017review}. Tensor-based methods such as TFN~\cite{zadeh2017tensor} model high-order inter-modal interactions, while sequence-level approaches like MFN~\cite{zadeh2018memory} leverage memory and attention mechanisms for temporal modeling.
The advent of transformer architectures marked a paradigm shift in MER. MulT~\cite{tsai2019multimodal} introduced cross-modal attention to capture long-range dependencies across modalities. MISA~\cite{hazarika2020misa} proposed disentangled representation learning to separate modality-invariant and 
modality-specific features. More recent approaches incorporate contrastive learning~\cite{zou2023multimodal} and multi-task learning~\cite{hu2022unimse} to further enhance multimodal representations.

Most recently, multimodal large language models (MLLMs) such as Video-LLaMA~\cite{zhang2023video}, PandaGPT~\cite{su2023pandagpt}, and Qwen2.5-Omni~\cite{xu2025qwen2} have demonstrated strong cross-modal perception by unifying text, audio, and visual understanding within a single architecture. Several studies have explored MLLMs for affective computing tasks~\cite{li2025multimodal,yang2024large,lian2024merbench}, showing promising zero-shot and few-shot performance. However, most existing methods, whether fusion-based or LLM-based, are designed for high-resource settings with abundant labeled data. Their effectiveness degrades substantially when applied to low-resource scenarios characterized by data scarcity, class imbalance, and 
cross-modal inconsistency.

\subsection{Low-Resource Multimodal Learning}
Low-resource multimodal learning presents challenges beyond data scarcity, including the lack of reliable preprocessing tools, pretrained models, and domain-specific linguistic resources~\cite{yan2023neural, ranathunga2023neural}. These challenges are amplified in multimodal contexts where aligning heterogeneous data across modalities introduces additional complexity~\cite{baltruvsaitis2018multimodal}.
Recent multilingual efforts have improved cross-lingual coverage in MER. Datasets such as MOUD~\cite{perez2013utterance}, CH-SIMS~\cite{yu2020ch}, and CMU-MOSEAS~\cite{zadeh2020cmu} extend coverage beyond English to Spanish, Chinese, Portuguese, German, and French. However, these primarily target 
high- to mid-resource languages. Southeast Asian languages, encompassing over 700 million speakers, remain severely underrepresented in multimodal affective computing research~\cite{magueresse2020low}.

To address data scarcity, recent work has explored transfer learning and cross-lingual adaptation by leveraging high-resource language models~\cite{ansell2021mad,yan2024neural, chen2025cross}. However, these approaches face 
fundamental limitations: morphological and syntactic mismatches across language families, culturally specific emotional expressions cannot be transferred, and a unified NLP infrastructure for many low-resource languages still lacks~\cite{ntalampiras2020toward}. These observations motivate the need for both dedicated datasets and specialized learning strategies tailored to low-resource multimodal emotion recognition.

\subsection{Auxiliary-task Learning}
Auxiliary-task learning provides additional supervisory signals that enhance representation learning~\cite{li2020multi}, offering an effective strategy for adapting large pretrained models to specialized tasks. In multimodal emotion recognition, auxiliary tasks have been designed for individual 
modalities to capture emotion-relevant features. For the textual modality, word-level polarity annotations help models identify emotion-critical vocabulary~\cite{shi2025mmkt}. For the visual modality, frame-level facial expression recognition guides the learning of emotion-aware visual representations~\cite{zheng2023facial}. For the acoustic modality, speech attributes such as pitch, speaking rate, and tone can be converted into natural language descriptions to enable LLM-based reasoning~\cite{wu2025beyond}. Additionally, hybrid strategies combining joint and multi-stage auxiliary learning have proven to be effective in conversational emotion recognition~\cite{shen2025coe,he2025dialoguemmt}.

Despite these advances, most existing auxiliary-task approaches are designed for high-resource settings and primarily focus on enhancing a single modality in isolation. In low-resource multimodal scenarios, where data scarcity coexists with cross-modal inconsistency, models are prone to learning spurious correlations from any single modality. A unified auxiliary learning strategy that jointly strengthens emotional representations across all modalities before fusion remains underexplored. This gap motivates our proposed OmniMER framework, which employs complementary auxiliary tasks across text, audio and video to improve unimodal emotional grounding before multimodal fusion.

\section{IndoMER Dataset}
\subsection{Data Collection}
\begin{figure}[htbp]
  \centering
  \includegraphics[width=0.8\linewidth]{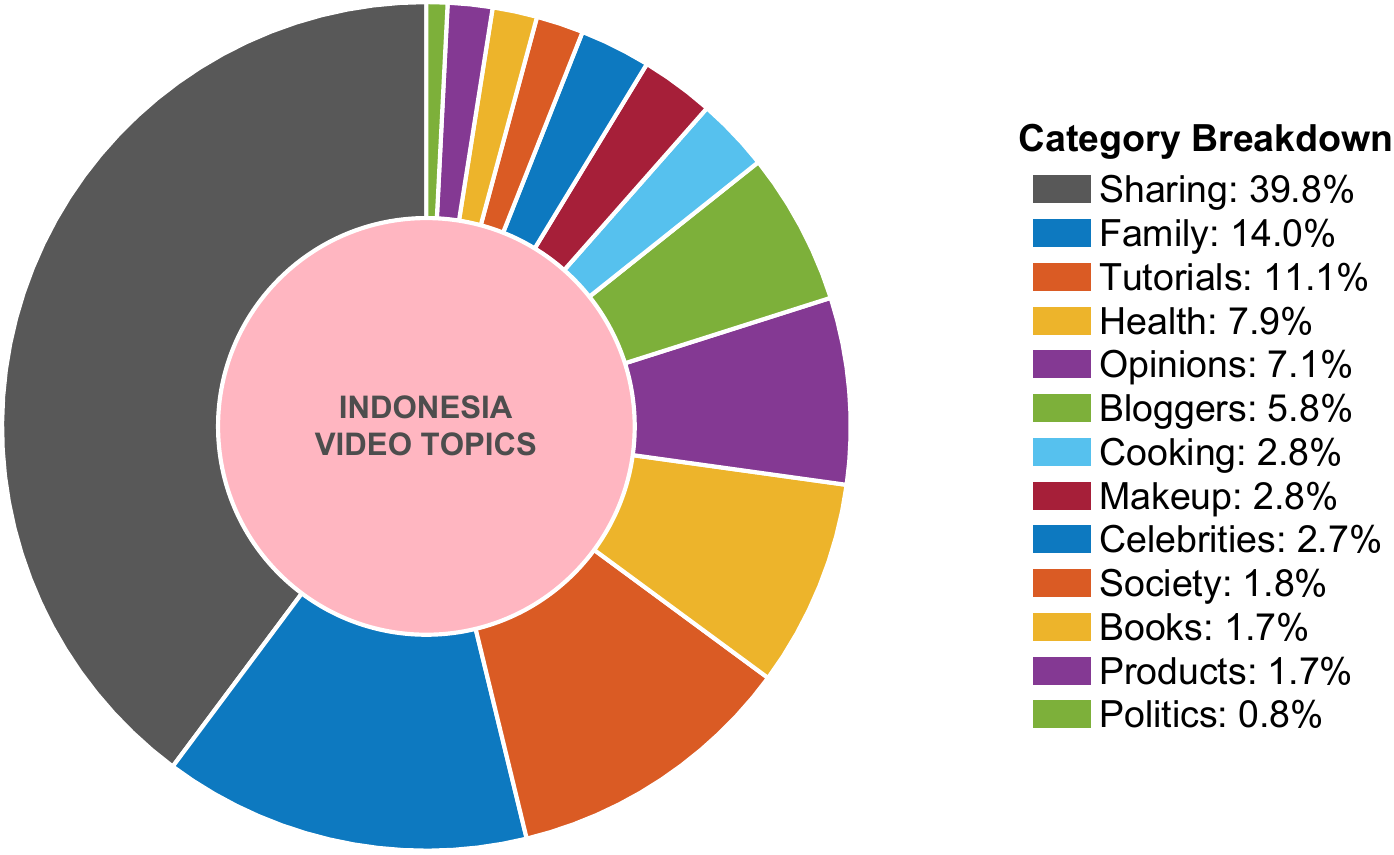}
  \caption{Categorical breakdown of video themes within the IndoMER dataset. The dataset spans thirteen distinct topics, ranging from personal life sharing to specific domains like health and politics. This wide variety of themes is selected to capture rich emotional nuances in different contexts.}
  \label{label}
\end{figure}

We collected individual monologue videos from publicly available social media platforms, such as YouTube and TikTok, prioritizing natural discourse where speakers show emotional expressions through language, vocal prosody, and facial dynamics. To ensure topic diversity and mitigate potential biases from single-domain focus, we conducted systematic searches covering 13 themes: bloggers, books, celebrities, cooking, family, health, makeup, personal opinions, politics (non-controversial discussions), products, sharing, society, and tutorials (Fig.~\ref{label}).

We applied the following selection criteria to maintain data quality and ethical standards: (1) each video contains only a single primary speaker to ensure clear modality alignment; (2) videos avoid sensitive topics 
related to religion, race, violence, or other controversial subjects; (3) content complies with platform community guidelines and contains no harmful, offensive, or politically inflammatory material; and (4) all 
videos are publicly available without violation of privacy or intellectual property rights.
Following these criteria, we obtained 207 source videos from 203 different speakers, ensuring both speaker diversity and data integrity. The gender distribution is approximately 40\% male and 60\% female, reflecting natural variations in creators social media contents.

\subsection{Segmentation and Annotation}

Segmentation and annotation were conducted by seven Indonesian speakers and one Indonesian language expert to ensure linguistic precision and cultural relevance. Videos were segmented based on speaker pauses and content boundaries. For each segment, manual transcription was performed with careful preservation of grammatical features and word usage patterns of natural speech, rather than normalizing to formal written Indonesian. The language expert reviewed all transcriptions for accuracy and consistency, with special attention to regional linguistic variations.

The annotation process was implemented in two phases. In the first phase, sentiment annotations were assigned to each segment on a scale from $-1$ (negative) to $+1$ (positive), with $0$ representing neutral. Each segment was independently labeled by five Indonesian speakers and the final sentiment label was determined via majority voting. 
In cases of disagreement, the language expert adjudicated based on cross-modal evidence, considering both linguistic content and non-verbal cues such as tone and facial expressions. In the second phase, emotion annotations were conducted using Ekman's six basic emotions plus a neutral category, yielding seven categories: fear, disgust, anger, sadness, happiness, surprise, and neutral. Each segment received independent 
ratings from five Indonesian speakers and one language expert using a Likert scale from 0 to 3 (0 = no sign, 1 = weak, 2 = moderate, 3 = strong). The final emotion label was determined by the highest cumulative score across annotators. In ambiguous cases, the language expert resolved disagreements by considering contextual and emotional nuances.

For annotation quality, we computed the percentage of segments where the majority vote was unambiguous (i.e., at least three annotators agreed on the same label). This rate was 78.1\% for sentiment and 82.3\% for emotion annotations. Segments with ambiguous votes were resolved by the language expert based on cross-modal 
evidence. 

\subsection{Dataset Statistics}
As summarized in Table~\ref{statistics_of_MIED}, the IndoMER dataset comprises 1,944 video segments from 207 source videos, representing 203 distinct speakers. The segments are concise and conversational, with an average duration of 5.37 seconds and an average of 11.68 words per segment. The content spans 13 distinct 
topics, reflecting a wide range of real-world emotional expressions on social media.

\begin{figure}[htbp]
  \centering 
  \includegraphics[width=\linewidth]{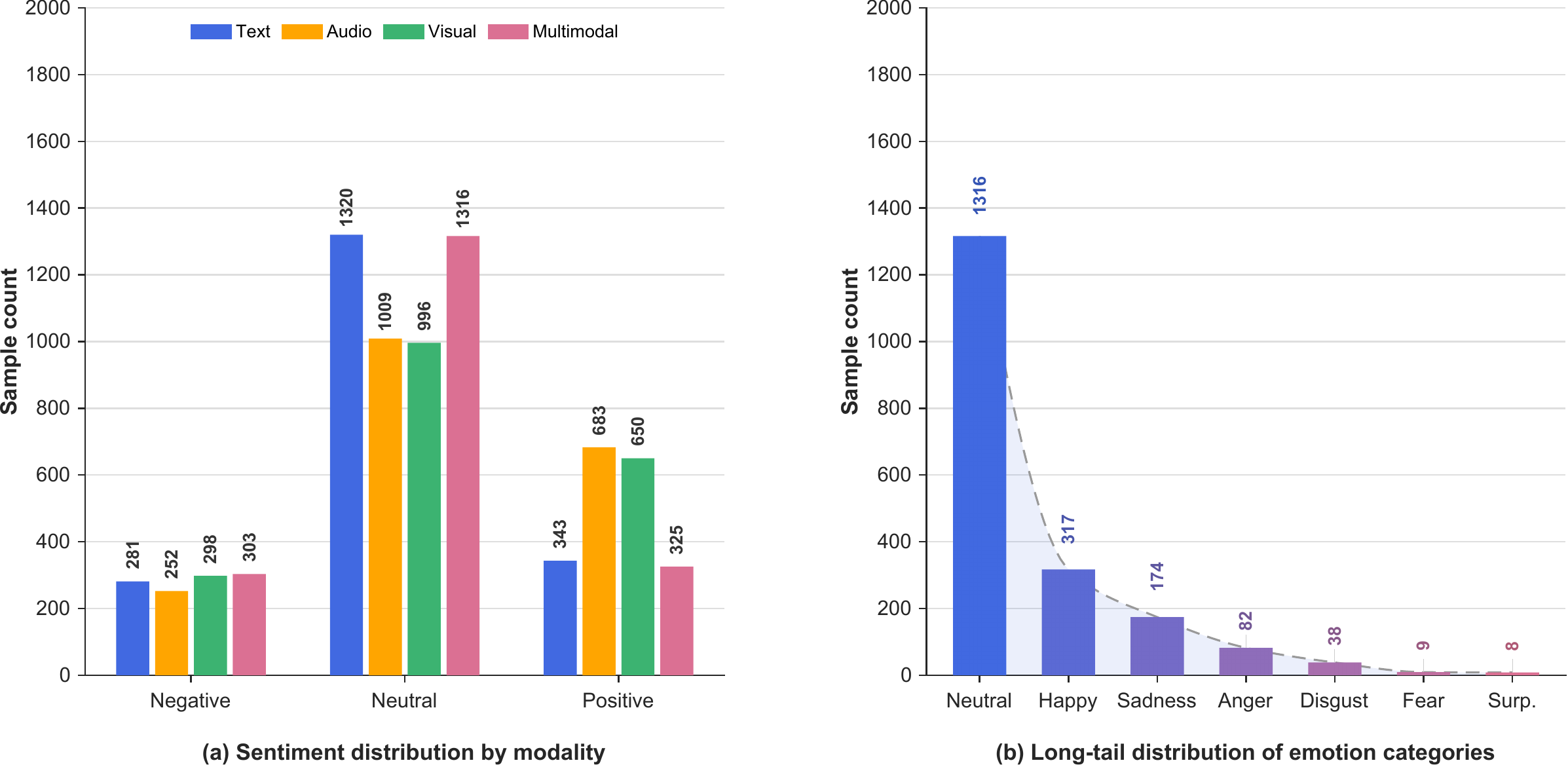} 
  \caption{Distribution of emotional and sentiment annotations in the IndoMER dataset. (a) Sentiment label distributions across text, audio, and visual modalities, compared with the multimodal consensus (ground truth). (b) Detailed breakdown of emotion categories within the multimodal annotations, highlighting the long-tailed nature of natural emotional communication.}
  \label{distribution_plot}
\end{figure}

Figure~\ref{distribution_plot} presents the distribution of sentiment and emotion annotations. The left panel (Fig.~\ref{distribution_plot}a) displays sentiment distributions for text, audio, and visual modalities individually, as well as the multimodal consensus. Notably, the multimodal distribution closely resembles 
the text modality distribution, indicating that textual information strongly influences overall emotional interpretation. This aligns with prior work showing that text often serves as the primary driver in multimodal sentiment analysis~\cite{deng2021survey,yan2022multimodal}. However, the divergence between text and audio/visual distributions on certain segments reveals that prosody and facial expressions provide complementary cues not fully captured by text alone.
The right panel (Fig.~\ref{distribution_plot}b) shows the emotion category distribution. The distribution exhibits a pronounced long-tailed pattern: neutral and happiness are substantially overrepresented, while fear and disgust are rare. This imbalance reflects natural emotional communication patterns but poses challenges for model training.

\begin{table}[htbp]
\centering
\caption{Key statistics of the IndoMER dataset.}
\label{statistics_of_MIED}
\begin{tabular}{lr}
\hline 
\textbf{Statistic} & \textbf{Value} \\
\hline 
Total source videos & 207 \\
Total video segments & 1,944 \\
Total distinct speakers & 203 \\
\quad \textit{- Male segments} & 778 \\
\quad \textit{- Female segments} & 1,167 \\
Average segment duration & 5.37 sec \\
Average word count & 11.68 words \\
Speech rate & 2.17 words/sec \\
Vocabulary size (unique words) & 4,066 \\ 
\hline 
\end{tabular}
\end{table}

\subsection{Feature Extraction}
To facilitate research using traditional multimodal fusion methods, we provide pre-extracted features for each modality following established practices in the MER literature.
For the textual modality, we employ IndoBERT\footnote{https://huggingface.co/indobenchmark/indobert-base-p1}, 
a pre-trained language model specifically trained on Indonesian corpora. Token sequences are standardized to a fixed length of 69 tokens through padding or truncation, yielding 768-dimensional embeddings per token.
For the acoustic modality, we utilize OpenSMILE~\cite{eyben2010opensmile} with the Geneva Minimalistic Acoustic Parameter Set (GeMAPS)~\cite{eyben2015geneva} to extract 18-dimensional Low-Level Descriptors (LLDs) from audio sampled at 22,050 Hz. These features capture prosodic information including frequency, energy and spectral characteristics. Acoustic sequences are standardized to 2,252 frames.
For the visual modality, we employ OpenFace 2.2\footnote{https://github.com/TadasBaltrusaitis/OpenFace} 
to extract 673-dimensional features from video frames at 25 fps, including facial action units, 2D/3D landmarks, head pose, and eye gaze. Visual sequences are standardized to 723 frames.

Note that the proposed OmniMER framework directly processes raw text, audio, and video inputs through Qwen2.5-Omni, which contains built-in encoders for each modality. The pre-extracted features described above are provided for researchers who wish to benchmark traditional fusion methods on IndoMER.

\begin{figure*}[ht!]
  \centering
  \includegraphics[width=0.95\linewidth]{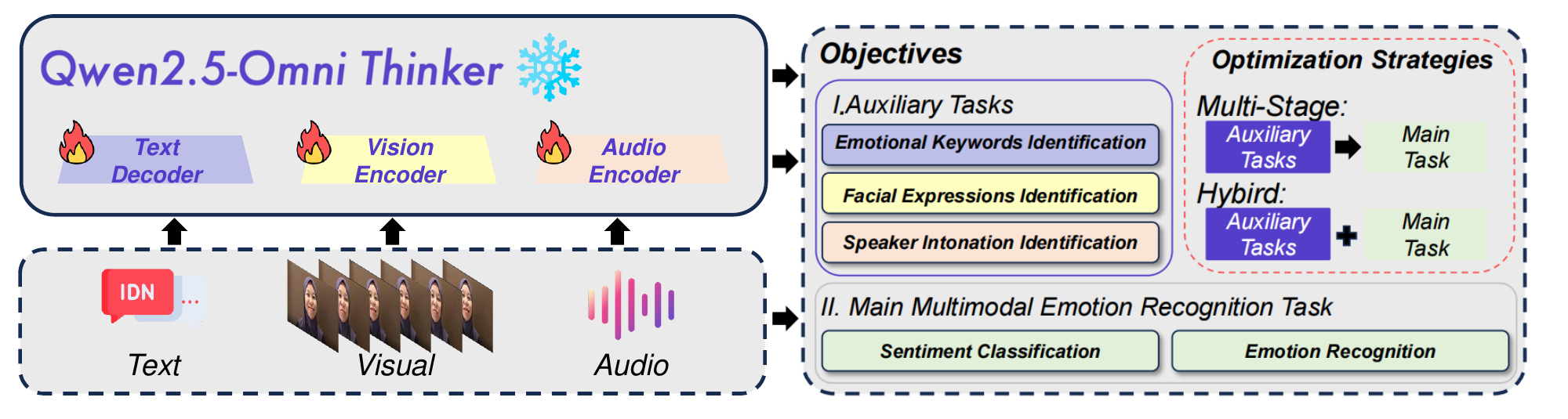}
  \caption{Overview of the proposed OmniMER framework. It leverages Qwen2.5-Omni as the unified multimodal backbone and incorporates modality-specific auxiliary tasks (text keywords, facial expressions, and speaker intonation) to enhance unimodal emotion grounding before multimodal fusion.}
  \label{fig:framework}
\end{figure*}

\subsection{Dataset Challenges}
The IndoMER dataset presents several challenges representative of low-resource multimodal emotion recognition. First, the dataset exhibits cross-modal inconsistency, where sentiment distributions vary across text, audio, and visual modalities (Fig.~\ref{distribution_plot}a). This requires models to resolve conflicting signals rather than relying on any single modality. Second, the emotion distribution follows a long-tailed pattern, with neutral and happiness dominating while fear and disgust are rare. This imbalance reflects Indonesian cultural norms that favor emotional restraint in public discourse, making the dataset culturally authentic but challenging for standard training procedures. Third, with 1,944 segments, the dataset is relatively small compared to high-resource benchmarks, increasing the risk of overfitting to spurious correlations. These characteristics motivate the need for specialized approaches that strengthen unimodal emotional grounding before multimodal fusion.

\section{OmniMER Framework}
Figure~\ref{fig:framework} illustrates the proposed OmniMER framework. Built upon Qwen2.5-Omni, an omni-modal LLM that supports text, audio, and visual inputs, OmniMER introduces modality-specific auxiliary perception tasks 
to strengthen unimodal emotional grounding prior to multimodal fusion. The key insight is that directly fusing multimodal features without ensuring the quality of unimodal representations can propagate noise and spurious correlations, especially in low-resource settings. 
To address this, OmniMER employs three auxiliary tasks that explicitly supervise emotion-relevant cue extraction in each modality: emotion keyword extraction for text, facial 
expression analysis for visual, and prosody analysis for audio. While the main task uses both sentiment and emotion labels, the auxiliary tasks operate under weaker supervision using only modality-level sentiment annotations. The supervision signals are automatically constructed by prompting the omni-modal backbone to generate modality-specific emotional interpretations, which are 
then validated against sentiment labels.
OmniMER consists of two core components: (1) modality-specific auxiliary perception tasks implemented via instruction-conditioned prompting, and (2) complementary training strategies, including multi-stage and hybrid optimization, to integrate auxiliary supervision with the main task.

\subsection{Main Multimodal Emotion Recognition Task}
In OmniMER, the main multimodal emotion recognition task is formulated as an instruction-conditioned generation problem. Given a multimodal input $x = (T, A, V)$ consisting of text, audio, and visual, the omni-modal LLM is prompted to produce structured emotional predictions.

We consider two task settings depending on label granularity. For coarse-grained sentiment classification, the model is instructed to directly predict sentiment polarity (positive, negative, or neutral) from the multimodal input. For fine-grained emotion recognition, we adopt a hierarchical formulation: the model first determines the sentiment polarity, then predicts a specific emotion category 
conditioned on the inferred sentiment. This hierarchical design leverages the observation that sentiment polarity provides a coarse emotional prior that constrains the space of possible emotions, reducing prediction ambiguity for fine-grained categories.

Formally, the main task is defined as
\begin{equation}
y = F_{\text{main}}(T, A, V; s_{\text{main}}),
\end{equation}
where $s_{\text{main}}$ denotes the task instruction, $F_{\text{main}}$ represents the instruction-conditioned behavior of the omni-modal LLM, and $y$ is the textual output specifying either sentiment polarity or a sentiment-conditioned emotion label.

For sentiment classification, the instruction prompts the model to analyze the speaker's overall emotional tone across all modalities and return a sentiment label. For emotion recognition, the instruction additionally requires the model to identify a specific emotion category from the seven predefined classes (fear, disgust, anger, sadness, happiness, surprise, and neutral) based on the inferred 
sentiment polarity.


\subsection{Auxiliary Tasks for Unimodal Emotional Grounding}

OmniMER introduces auxiliary perception tasks for text, video, and audio modalities to strengthen unimodal emotional grounding before fusion. Unlike previous auxiliary-task approaches that typically focus on a single modality or require additional manual annotations~\cite{shi2025mmkt,zheng2023facial}, our design has two distinguishing features. First, we employ a unified instruction-conditioned prompting paradigm across all three modalities, enabling consistent implementation within the omni-modal LLM framework. Second, the auxiliary supervision is automatically constructed using only existing modality-level sentiment labels, avoiding the need for fine-grained annotations such as word-level polarity or frame-level expressions.

Each auxiliary task is formulated as a modality-specific function that produces both a sentiment prediction and an interpretable explanation grounded in modality-specific evidence. The outputs are constrained to the structured JSON format to ensure consistency and facilitate automatic validation.


\subsubsection{Text Auxiliary Task: Emotional Keywords
Identification}

Given a tokenized Indonesian utterance $T = \{w_1, \dots, w_n\}$, the text that auxiliary task aims to ground sentiment inference in emotion-related lexical cues by identifying salient keywords and phrases in the utterance:
\begin{equation}
(W_{\text{text}}, S_{\text{text}}, EX_{\text{text}}) = F_{\text{text}}(T),
\end{equation}
where $W_{\text{text}}$ denotes the set of detected emotion-related keywords or phrases,
$S_{\text{text}}$ represents the implied sentiment polarity,
and $EX_{\text{text}}$ is a short natural-language explanation describing how the identified keywords provide evidence that grounds the inferred sentiment polarity.

The function $F_{\text{text}}$ is implemented via an instruction prompt that requires the model to make its emotion-related lexical evidence explicit when producing a sentiment judgment.
The output is constrained to a structured JSON format:

\begin{promptbox}{Prompt: Identification of Emotional Keywords }

\textbf{Instruction.} You are an assistant for sentiment analysis of Indonesian text.

\textbf{Input.} Text: \{\textit{text}\}

\textbf{Task.} Identify emotion-related keywords or phrases in the text and determine the overall sentiment supported by these cues.

\textbf{Return (JSON only).}
\begin{jsonbox}
"Sentiment": "positive | negative | neutral",
"Emotion_keywords": ["keyword1", "keyword2", "..."],
"Explanation": "A brief analysis grounded in the identified keywords."
\end{jsonbox}
\end{promptbox}

By explicitly requiring emotion-bearing lexical cues, this auxiliary task encourages the model to ground sentiment predictions in linguistically salient affective evidence rather than relying on superficial lexical or distributional correlations.

\subsubsection{Visual Auxiliary Task: Facial Expressions Identification}

A video clip is represented as a sequence of frames $V = \{I_1, \dots, I_m\}$.
For each frame, the model attends to visual expression cues such as facial muscle movements, gaze changes, and micro-expressions, as well as their temporal dynamics.
The visual auxiliary task aims to ground sentiment inference in facial dynamics and provide a concise visually grounded interpretation:
\begin{equation}
(S_{\text{visual}}, EX_{\text{visual}}) = F_{\text{visual}}(V),
\end{equation}
where $S_{\text{visual}}$ denotes the implied sentiment polarity inferred from visual expressions, and $EX_{\text{visual}}$ is a short natural-language explanation grounded in observable facial behaviors and their temporal dynamics.

The function $F_{\text{visual}}$ is implemented via an instruction prompt that guides the model to make explicit use of facial expressions and their temporal variations when producing a sentiment judgment.
The output is constrained to a structured JSON format to ensure consistency and interpretability:

\begin{promptbox}{Prompt: Facial Expressions Identification}

\textbf{Instruction.} You are an assistant for sentiment analysis of a visual clip.

\textbf{Input.} Visual: \{\textit{visual}\}

\textbf{Task.} Analyze the speaker’s facial expressions and their temporal variations, and determine the overall sentiment supported by these visual cues.

\textbf{Return (JSON only).}
\begin{jsonbox}
"Sentiment": "positive | negative | neutral",
"Explanation": "A brief analysis grounded in observable facial expressions and their temporal dynamics."
\end{jsonbox}

\end{promptbox}

By explicitly focusing on facial expressions and temporal visual cues, this auxiliary task encourages the model to ground affective inference in concrete visual evidence rather than relying on textual or contextual biases.

\subsubsection{Audio Auxiliary Task: Speaker Intonation Identification}

The audio signal is segmented into short fragments $A = \{A_1, \dots, A_k\}$.
For each fragment, the model attends to acoustic cues such as intonation, pitch variation, loudness, and speaking rate, as well as their temporal patterns.
The audio auxiliary task aims to ground sentiment inference in speech prosody and provide a concise acoustically grounded interpretation:
\begin{equation}
(S_{\text{audio}}, EX_{\text{audio}}) = F_{\text{audio}}(A),
\end{equation}
where $S_{\text{audio}}$ denotes the implied sentiment polarity inferred from prosodic patterns, and $EX_{\text{audio}}$ is a short natural-language explanation describing how acoustic cues provide evidence that grounds the inferred sentiment.

The function $F_{\text{audio}}$ is implemented via an instruction prompt that guides the model to make explicit use of prosodic variations and speaking style when producing a sentiment judgment.
The output is constrained to a structured JSON format:

\begin{promptbox}{Prompt: Speaker Intonation Identification}

\textbf{Instruction.} You are an assistant for sentiment analysis of an audio clip.

\textbf{Input.} Audio: \{\textit{audio}\}

\textbf{Task.} Analyze prosodic cues such as intonation, speaking rate, and vocal dynamics, and determine the overall sentiment supported by these acoustic patterns.

\textbf{Return (JSON only).}
\begin{jsonbox}
"Sentiment": "positive | negative | neutral",
"Explanation": "A brief analysis grounded in prosodic cues such as intonation and speaking rate."
\end{jsonbox}

\end{promptbox}

By emphasizing prosodic and temporal speech cues, this auxiliary task encourages the model to ground affective inference in acoustic evidence conveyed through vocal delivery, complementing textual and visual emotional cues.

\subsection{Auxiliary Supervision Construction}
\begin{figure}[t]
  \centering
  \includegraphics[width=0.9\linewidth]{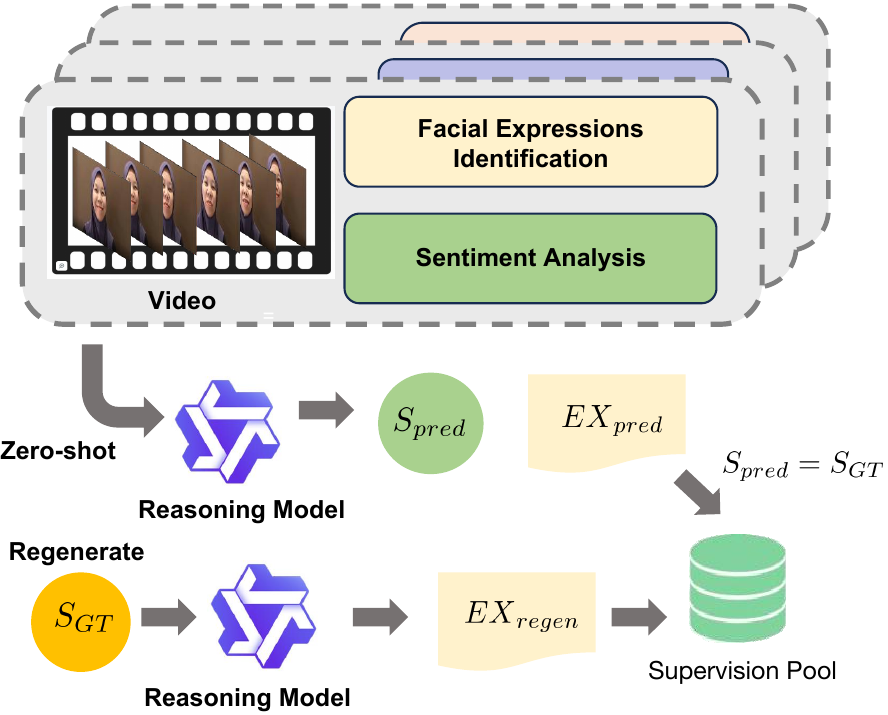}
  \caption{Construction of modality-specific auxiliary supervision from sentiment labels. Explanations are retained only when their associated sentiment predictions are consistent with ground-truth annotations, or regenerated under sentiment constraints otherwise.}
  \label{fig:aux_construct}
\end{figure}

Although IndoMER provides sentiment annotations at both multimodal and modality-specific levels, it does not include fine-grained annotations describing modality-specific emotional evidence, such as emotion keywords in text, facial expressions in visual or prosodic cues in audio. To obtain supervision for the auxiliary tasks, we leverage the generative capabilities of Qwen2.5-Omni with a two-step consistency-check mechanism (Fig.~\ref{fig:aux_construct}).

For each modality $m \in \{\text{text}, \text{visual}, \text{audio}\}$, the model is first prompted to produce a sentiment prediction and a corresponding explanation based on the modality input $x^{(m)}$:
\begin{equation}
\left(S_{\text{pred}}^{(m)}, EX_{\text{pred}}^{(m)}\right) = \mathcal{G}\!\left(x^{(m)}\right),
\end{equation}
where $\mathcal{G}(\cdot)$ denotes instruction-conditioned generation.

The predicted sentiment $S_{\text{pred}}^{(m)}$ is compared with the ground-truth label $s_{\text{GT}}^{(m)}$. If they match, the generated explanation $EX_{\text{pred}}^{(m)}$ is retained as valid supervision. Otherwise, the model is re-invoked with the ground-truth sentiment as a constraint:
\begin{equation}
EX^{(m)} = \mathcal{G}\!\left(x^{(m)} \mid s_{\text{GT}}^{(m)}\right),
\end{equation}
In this case, the model generates an explanation that justifies the given sentiment label rather than re-predicting sentiment.

This mechanism ensures that auxiliary explanations are either self-consistent with the model's prediction or explicitly aligned with ground-truth annotations, yielding reliable supervision without requiring manual annotation of modality-specific emotional evidence.

\subsection{Training and Optimization Strategies}


OmniMER adapts Qwen2.5-Omni through parameter-efficient fine-tuning using Low-Rank Adaptation (LoRA). We insert LoRA modules into the self-attention layers of all modality encoders (text, audio, and visual) as well as the multimodal Thinker responsible for cross-modal reasoning. All backbone parameters are frozen, and only the LoRA parameters are updated during training. This design allows auxiliary supervision and the main task to jointly guide the adaptation of attention mechanisms, enabling the model to emphasize emotion-relevant cues within each modality while preserving the pretrained multimodal representations.

\subsubsection{Unified Instruction-Conditioned Objective}
All tasks are formulated under a unified instruction-conditioned generation paradigm. Let $\mathcal{D}=\{(x_i, s_i, y_i)\}_{i=1}^{N}$ denote the training set, where $x_i=(T_i, A_i, V_i)$ is the multimodal input, $s_i$ is a task instruction, and $y_i$ is the target output (e.g., a sentiment label, an emotion label, or a modality-specific explanation). Training minimizes the negative log-likelihood:
\begin{equation}
\mathcal{L}_{\text{gen}}(\theta) = -\sum_{i=1}^{N}\log p_{\theta}\!\left(y_i \mid x_i, s_i\right),
\label{eq:gen_loss}
\end{equation}
where $\theta$ denotes the trainable LoRA parameters. This unified objective enables auxiliary supervision and the main task to share the same learning signal without requiring task-specific prediction heads.

\subsubsection{Instruction Scheduling Strategies}

Although all tasks share the same objective, they differ in how instructions are scheduled during optimization. We define instruction types $\mathcal{S}=\{\textsc{Aux-Text}, \textsc{Aux-Audio}, \textsc{Aux-Visual}, \textsc{Main}\}$ and explore two scheduling strategies.

The first is multi-stage optimization, which decouples unimodal grounding from target prediction using a two-stage schedule:
\begin{equation}
\pi_t(s)=
\begin{cases}
\pi_{\text{aux}}(s), & t \leq t_0,\\
\pi_{\text{main}}(s), & t > t_0,
\end{cases}
\label{eq:stage_schedule}
\end{equation}
where $\pi_{\text{aux}}$ samples only auxiliary instructions and $\pi_{\text{main}}$ samples only the main task instruction. Stage I strengthens modality-specific emotional understanding, while Stage II fine-tunes the model on the target task using the grounded representations learned during Stage I.

The second is hybrid mixed-instruction optimization, which performs single-stage training by sampling from all instruction types throughout:
\begin{equation}
\pi_t(s) = \pi_{\text{mix}}(s),  \forall\,t,
\label{eq:hybrid_schedule}
\end{equation}
where $\pi_{\text{mix}}$ draws from both auxiliary and main tasks. This strategy encourages early interaction between unimodal grounding and multimodal reasoning.

Both strategies optimize the same objective in Eq.~(\ref{eq:gen_loss}) and differ only in how auxiliary supervision is integrated. As shown in our experiments, hybrid optimization achieves stronger performance on sentiment classification, while multi-stage optimization provides more stable gains for emotion recognition under long-tailed distributions.

\section{Experiments}
This section evaluates the proposed OmniMER framework on the IndoMER dataset 
for two tasks: multimodal sentiment classification and fine-grained emotion 
recognition. We analyze the impact of modality-specific auxiliary tasks under 
class imbalance and cross-modal inconsistency, and further examine cross-lingual 
generalization on the CH-SIMS dataset.

\subsection{Experimental Setup}
All experiments are conducted on the IndoMER dataset, which contains 1,944 
temporally aligned text-audio-visual samples. Two tasks are evaluated: sentiment 
classification (negative, neutral, positive) and emotion recognition with seven 
categories. Macro-F1 is used as the primary metric due to severe class imbalance, 
with Accuracy and Weighted-F1 as supplementary metrics.
We adopt Qwen2.5-Omni-7B as the backbone and perform parameter-efficient 
fine-tuning using LoRA. LoRA modules are applied to the attention layers of 
all modality encoders and the multimodal Thinker, with rank set to 8 and 
scaling factor $\alpha=16$. The vision tower and multimodal projector are frozen.

Training is performed on two NVIDIA RTX 3090 GPUs. We use the AdamW optimizer 
with an initial learning rate of $5\times10^{-5}$ and cosine annealing. The 
batch size is 1 per GPU with 64 gradient accumulation steps (effective batch 
size = 128). All models are trained for 5 epochs.

\subsection{Main Results}
Tables~\ref{tab:sentiment_main} and \ref{tab:emotion_main} report the main 
results on sentiment classification and emotion recognition. Due to severe 
class imbalance and cross-modal inconsistency, the Base model that trained without 
auxiliary supervision achieves a Macro-F1 of only 0.506 for sentiment 
classification and 0.233 for emotion recognition.
Introducing a single auxiliary task consistently improves Macro-F1 across 
both tasks. Among the three modalities, the video auxiliary task yields the 
most noticeable gains, indicating that facial expressions provide relatively 
direct and discriminative affective cues. Nevertheless, improvements from 
any single auxiliary modality remain limited, suggesting that unimodal 
guidance alone is insufficient to address multimodal ambiguity and long-tailed 
distributions.

Jointly integrating all auxiliary tasks leads to substantial performance gains. 
For sentiment classification, the Hybrid strategy achieves the best performance, 
improving Macro-F1 from 0.506 to 0.582 (an absolute gain of 7.6 points). This 
indicates that hybrid mixed-instruction training effectively enhances 
discriminative capability, particularly for common sentiment categories. For 
emotion recognition, the Multi-stage strategy yields the most significant 
improvement, raising Macro-F1 from 0.233 to 0.454 (an absolute gain of 22.1 
points). This highlights the advantage of decoupling unimodal grounding from 
target prediction when dealing with fine-grained, long-tailed emotion classes.

Overall, these results demonstrate that auxiliary modality-specific tasks are 
complementary rather than redundant. Hybrid training favors coarse-grained 
sentiment discrimination, while the multi-stage strategy is more effective 
for fine-grained emotion recognition under class-imbalanced conditions.

\begin{table}[t]
\centering
\caption{Main results on the IndoMER sentiment classification task.}
\label{tab:sentiment_main}
\renewcommand{\arraystretch}{1.15}
\begin{tabular}{lccc}
\toprule
\textbf{Model} & \textbf{Accuracy} & \textbf{Macro F1} & \textbf{Weighted F1} \\
\midrule
Base                         & 0.702 & 0.506 & 0.693 \\
Base + Text Aux              & 0.730 & 0.516 & 0.711 \\
Base + Video Aux             & 0.725 & 0.530 & 0.709 \\
Base + Audio Aux             & 0.733 & 0.518 & 0.714 \\
\midrule
All Aux (Multi-stage)        & \underline{0.742} & \underline{0.536} & \underline{0.724} \\
All Aux (Hybrid)             & \textbf{0.744} & \textbf{0.582} & \textbf{0.739} \\
\bottomrule
\end{tabular}
\end{table}

\begin{table}[t]
\centering
\caption{Main results on the IndoMER emotion recognition task.}
\label{tab:emotion_main}
\renewcommand{\arraystretch}{1.15}
\begin{tabular}{lccc}
\toprule
\textbf{Model} & \textbf{Accuracy} & \textbf{Macro F1} & \textbf{Weighted F1} \\
\midrule
Base                         & 0.694 & 0.233 & 0.669 \\
Base + Text Aux              & 0.711 & 0.238 & 0.667 \\
Base + Visual Aux             & \underline{0.730} & 0.291 & \underline{0.694} \\
Base + Audio Aux             & 0.714 & 0.271 & 0.675 \\
\midrule
All Aux (Multi-stage)        & 0.711 & \textbf{0.454} & \textbf{0.706} \\
All Aux (Hybrid)             & \textbf{0.739} & \underline{0.315} & 0.693 \\
\bottomrule
\end{tabular}
\end{table}

\begin{table*}[t]
\centering
\caption{Sentiment-wise performance on the IndoMER dataset.}
\label{tab:Sentiment_wise}
\renewcommand{\arraystretch}{1.15}
\begin{tabular}{lcccccc}
\toprule
\textbf{Class} 
& Base 
& Base + Text Aux 
& Base + Visual Aux 
& Base + Audio Aux 
& All Aux (Multi-stage) 
& All Aux (Hybrid) \\
\midrule
Neutral  
& 0.832 
& 0.837 
& 0.827 
& 0.839 
& \underline{0.846} 
& \textbf{0.859} \\

Positive 
& 0.308 
& 0.323 
& \underline{0.387} 
& 0.300 
& 0.358 
& \textbf{0.466} \\

Negative 
& 0.396 
& 0.388 
& 0.376 
& \underline{0.416} 
& 0.404 
& \textbf{0.438} \\
\bottomrule
\end{tabular}
\end{table*}

\begin{table*}[t]
\centering
\caption{Emotion-wise performance on the IndoMER dataset.}
\label{tab:Emotion_wise}
\renewcommand{\arraystretch}{1.15}
\begin{tabular}{lcccccc}
\toprule
\textbf{Class} 
& Base 
& Base + Text Aux 
& Base + Visual Aux 
& Base + Audio Aux 
& All Aux (Multi-stage) 
& All Aux (Hybrid) \\
\midrule
Fear (ketakutan) 
& 0.133 
& 0.250 
& 0.286 
& \textbf{0.400} 
& \underline{0.307} 
& 0.286 \\

Disgust (jijik) 
& 0 
& 0 
& 0 
& 0 
& 0 
& \textbf{0.286} \\

Anger (kemarahan) 
& 0.154 
& 0.087 
& \textbf{0.261} 
& 0.095 
& 0 
& \underline{0.240} \\

Sadness (kesedihan) 
& 0.240 
& 0.333 
& 0.327 
& 0.340 
& \textbf{0.379} 
& \underline{0.375} \\

Neutral (netral) 
& 0.827 
& 0.834 
& 0.837 
& 0.830 
& 0.832 
& \textbf{0.848} \\

Happiness (kebahagiaan) 
& 0.276 
& 0.163 
& \underline{0.327} 
& 0.235 
& \textbf{0.400} 
& 0.174 \\

Surprise 
& 0 
& 0 
& 0 
& 0 
& \textbf{1.000} 
& 0 \\
\bottomrule
\end{tabular}
\end{table*}

\subsection{Sentiment- and Emotion-wise Performance Analysis}
A fine-grained sentiment- and emotion-wise F1 analysis 
(Tables~\ref{tab:Sentiment_wise} and \ref{tab:Emotion_wise}) provides deeper 
insights into how auxiliary tasks reshape model behavior across affective 
categories. Beyond improving overall performance, auxiliary supervision 
substantially enhances recognition of minority classes under IndoMER's 
long-tailed distribution.

For sentiment classification, all models achieve strong performance on the 
majority \textit{neutral} class, with the hybrid strategy attaining the highest 
F1 score of 0.859. More pronounced gains are observed for minority categories. 
The \textit{positive} class benefits most from auxiliary supervision, with its 
F1 score increasing from 0.308 (Base) to 0.466 under the hybrid strategy, 
constituting the largest improvement among sentiment labels. The \textit{negative} 
class also shows consistent improvement, rising from 0.396 to 0.438. These 
results indicate that auxiliary tasks significantly enhance discriminative 
capability for low-frequency sentiment categories.

For emotion recognition, the impact of auxiliary learning is even more evident. 
Both hybrid and multi-stage strategies recover several low-frequency emotions 
that the Base model fails to recognize. Notably, the \textit{surprise} class, 
unrecognized by the Base and all single-auxiliary models, achieves perfect 
recognition (F1 = 1.000) under the multi-stage strategy. The \textit{happiness} 
and \textit{sadness} classes show substantial gains under the multi-stage 
strategy, with F1 scores improving from 0.276 to 0.400 and from 0.240 to 0.379, 
respectively. For modality-sensitive emotions, single-modality auxiliary tasks 
are also effective: the audio auxiliary task boosts \textit{fear} from 0.133 to 
0.400, while the video auxiliary task improves \textit{anger} from 0.154 to 0.261. 
The majority \textit{neutral} class benefits most from the hybrid strategy, 
increasing from 0.827 to 0.848.

Overall, these class-wise results demonstrate that auxiliary tasks complement 
each other by strengthening unimodal emotional perception. The multi-stage 
strategy is particularly effective for recovering low-frequency emotions, 
whereas the hybrid strategy provides more stable improvements for majority 
categories. Despite these gains, recognizing extremely rare emotions such as 
\textit{disgust} remains challenging, underscoring the difficulty of long-tailed 
recognition under limited supervision.

\subsection{Case Study}
\begin{figure}[t]
   \centering
   \includegraphics[width=\linewidth]{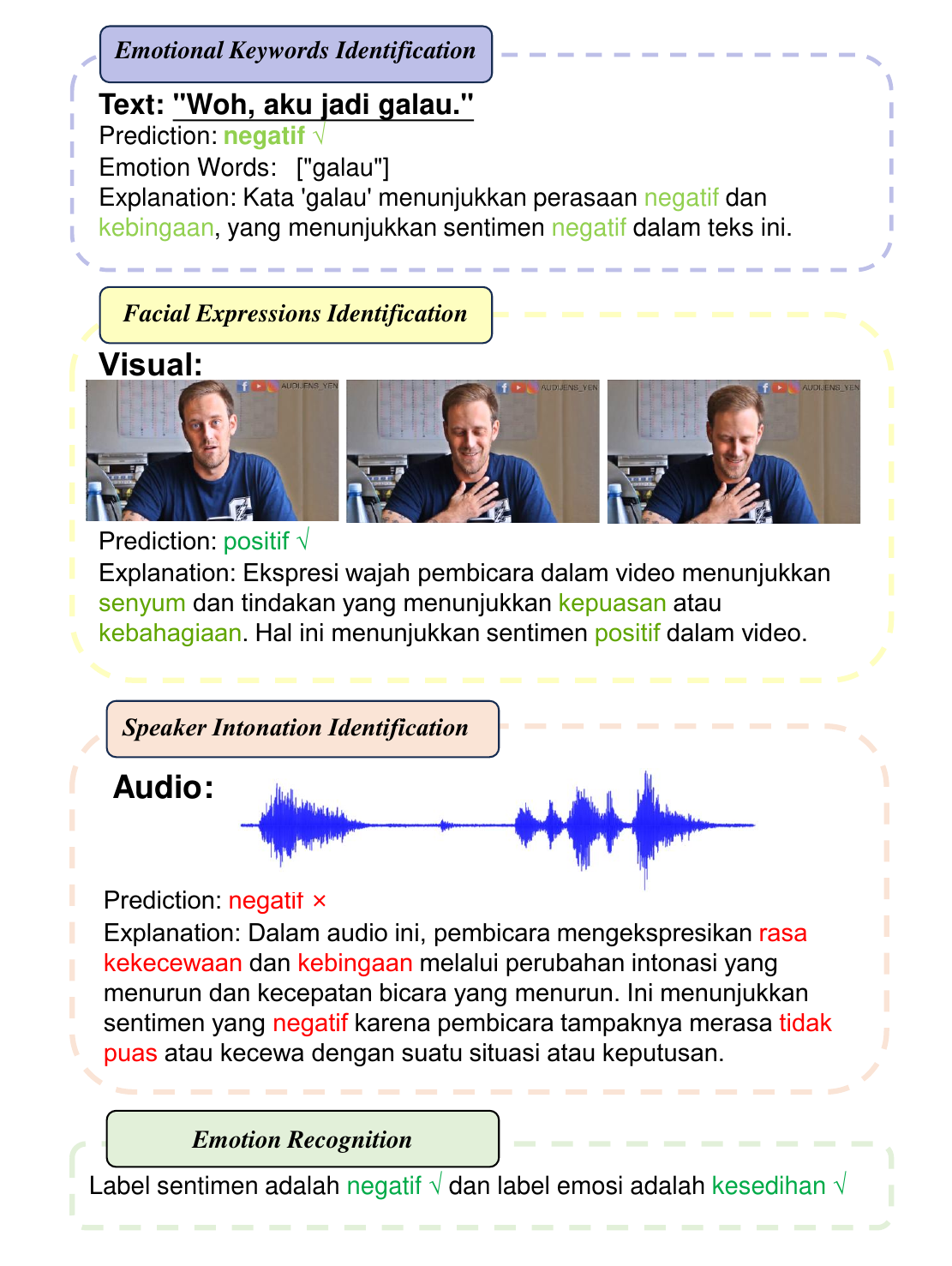}
   \caption{A representative example illustrating how modality-specific auxiliary tasks expose complementary and conflicting emotional cues across text, video, and audio, and how OmniMER integrates these cues to produce a robust multimodal emotion prediction.}
   \label{fig:a_specific_example}
\end{figure}

To qualitatively illustrate how modality-specific auxiliary tasks support multimodal emotion recognition, Fig.~\ref{fig:a_specific_example} presents a representative example from IndoMER that exhibits clear cross-modal inconsistency.

The textual utterance contains the Indonesian expression ``galau'', which explicitly conveys negative affect and emotional distress. The text auxiliary task correctly identifies this emotion-bearing keyword and provides a strong linguistic signal for negative polarity and sadness. In contrast, the visual stream shows a slight smile and relaxed facial movements, which could be misleading if considered in isolation. The video auxiliary task captures 
these facial cues and interprets them as a positive affect, revealing a modality-level conflict rather than suppressing it.

The audio auxiliary task analyzes prosodic patterns such as intonation and 
speaking rate, producing a negative-oriented interpretation. However, the 
acoustic evidence is relatively weak and ambiguous in this case, reflecting 
the inherent uncertainty of paralinguistic cues. Importantly, OmniMER does 
not require that all auxiliary tasks be individually correct. Instead, these 
modality-specific analyses collectively expose complementary and potentially 
conflicting emotional evidence.

By integrating these auxiliary outputs, the final multimodal prediction 
correctly identifies the sentiment as negative and the emotion as sadness. 
This example demonstrates that auxiliary tasks help disentangle modality 
conflicts by making unimodal emotional cues explicit, enabling more robust 
and interpretable multimodal reasoning even under ambiguous or conflicting 
signals.

\begin{table}
\centering
\caption{Results on the CH-SIMS sentiment classification task.}
\label{tab:sentiment_zh}
\renewcommand{\arraystretch}{1.15}
\begin{tabular}{lccc}
\toprule
\textbf{Model} & \textbf{Accuracy} & \textbf{Macro F1} & \textbf{Weighted F1} \\
\midrule
Base                         & 0.702 & 0.596 & 0.696 \\
Base + Text Aux              & 0.735 & 0.625 & 0.718 \\
Base + Visual Aux             & 0.722 & 0.609 & 0.722 \\
Base + Audio Aux             & 0.730 & 0.619 & 0.716 \\
\midrule
All Aux (Multi-stage)        & \underline{0.742} & \underline{0.628} & \underline{0.723} \\
All Aux (Hybrid)             & \textbf{0.751} & \textbf{0.652} & \textbf{0.738} \\
\bottomrule
\end{tabular}
\end{table}

\begin{table*}
\centering
\caption{Sentiment-wise performance on the CH-SIMS dataset.}
\label{tab:Sentiment_wise_zh}
\renewcommand{\arraystretch}{1.15}
\begin{tabular}{lcccccc}
\toprule
\textbf{Class} 
& Base 
& Base + Text Aux 
& Base + Visual Aux 
& Base + Audio Aux 
& All Aux (Multi-stage) 
& All Aux (Hybrid) \\
\midrule
Neutral  
& 0.281 
& \underline{0.315} 
& 0.278 
& 0.304 
& 0.302 
& \textbf{0.368} \\

Positive 
& 0.693 
& 0.745 
& 0.741 
& 0.735 
& \textbf{0.764} 
& \underline{0.758} \\

Negative 
& 0.813 
& 0.816 
& 0.809 
& \underline{0.819} 
& 0.818 
& \textbf{0.829} \\
\bottomrule
\end{tabular}
\end{table*}

\subsection{Cross-lingual Validation}
To evaluate cross-lingual generalization, we conduct experiments on the 
Chinese multimodal sentiment dataset CH-SIMS. CH-SIMS differs substantially 
from IndoMER in language structure, emotional expression patterns, and class 
distribution, providing a rigorous testbed for cross-lingual transfer. All 
experiments follow the same configuration and training protocol as IndoMER.

Table~\ref{tab:sentiment_zh} reports the overall sentiment classification 
results. The Base model achieves a Macro-F1 of 0.596 without auxiliary 
supervision. Introducing any single auxiliary task consistently improves 
Macro-F1, indicating that auxiliary tasks contribute to learning transferable 
affective cues across languages. When all auxiliary tasks are integrated, 
performance is further improved. The hybrid strategy achieves the best result 
with a Macro-F1 of 0.652, while the multi-stage strategy attains 0.628. These 
results demonstrate that auxiliary-task integration generalizes well beyond 
Indonesian to Chinese multimodal sentiment analysis.

Table~\ref{tab:Sentiment_wise_zh} presents sentiment-wise F1 scores. Notably, 
the class distribution in CH-SIMS differs from IndoMER: the \textit{Neutral} 
class is a minority category, whereas the \textit{Negative} class dominates. 
The Base model performs poorly on the minority \textit{Neutral} class with 
an F1 of only 0.281, while the hybrid strategy improves this to 0.368 (an 
absolute gain of 8.7 points). This confirms that auxiliary tasks benefit 
long-tailed categories across languages.
For the \textit{Positive} class, the multi-stage strategy achieves the highest 
F1 (0.764), suggesting that staged auxiliary learning is effective for 
capturing positive affect cues. For the majority \textit{Negative} class, all 
models perform strongly, with the hybrid strategy further improving F1 from 
0.813 to 0.829, indicating that auxiliary-task integration enhances 
majority-class performance without sacrificing minority-class recognition.

Overall, these results demonstrate that OmniMER captures language-agnostic 
emotional patterns and transfers effectively across languages with different 
distributions and expressive conventions, validating it as a practical 
framework for low-resource and cross-lingual multimodal emotion recognition.

\section{Conclusions}
In this study, we addressed multimodal emotion recognition in low-resource language settings. We introduced IndoMER, the first multimodal emotion recognition benchmark for Indonesian, comprising 1,944 video segments from 208 speakers with aligned text, audio, and visual annotations across seven 
emotion categories. IndoMER presents realistic challenges including cross-modal inconsistency and long-tailed class distributions that reflect Indonesian cultural communication norms.
To tackle these challenges, we proposed OmniMER, a framework that adapts omni-modal LLMs through auxiliary modality-specific perception tasks to strengthen unimodal emotional representations prior to fusion. Experiments on IndoMER show that OmniMER outperforms the base model in Macro-F1 on sentiment classification and emotion recognition, with notable gains on minority emotion categories. Cross-lingual evaluation on CH-SIMS further validates the generalizability of the framework.This work has limitations: the dataset remains relatively small, and experiments focus on a single backbone model. Future work includes extending to additional 
underrepresented languages, incorporating more baseline comparisons, and exploring finer-grained auxiliary objectives.





\bibliographystyle{IEEEtran}
\bibliography{IEEEabrv, Bibliography}




\end{document}